# Robotics for poultry farming: challenges and opportunities

Ugur Ozenturk[a,b,*], Zhengqi Chen[c], Lorenzo Jamone[c], Elisabetta Versace[a,*]

[a] School of Biological and Behavioural Sciences, Department of Biological and Experimental Psychology, Queen Mary University of London, UK; [b] Ataturk University, Faculty of Veterinary Medicine, Department of Animal Science, Turkiye; [c] ARQ (Advanced Robotics at Queen Mary), School of Engineering and Materials Science, Queen Mary University of London, UK

* Corresponding authors. Email address: ugur.ozenturk@atauni.edu.tr (Ugur Ozenturk); e.versace@qmul.ac.uk (Elisabetta Versace)

**Abstract**

Poultry farming plays a pivotal role in addressing human food demand. Robots are emerging as promising tools in poultry farming, with the potential to address sustainability issues while meeting the increasing production needs and demand for animal welfare. This review aims to identify the current advancements, limitations and future directions of development for robotics in poultry farming by examining existing challenges, solutions and innovative research, including robot-animal interactions. We cover the application of robots in different areas, from environmental monitoring to disease control, floor eggs collection and animal welfare. Robots not only demonstrate effective implementation on farms but also hold potential for ethological research on collective and social behaviour, which can in turn drive a better integration in industrial farming, with improved productivity and enhanced animal welfare.

# 1. Introduction

Poultry farming plays a crucial role in meeting the growing demand for affordable and safe food products (Sarıca et al., 2018). Poultry production is cost-effective (Ahmad et al., 2022; de Mesquita Souza Saraiva et al., 2022) and offers high-quality proteins (Attia et al., 2022; de Mesquita Souza Saraiva et al., 2022). Furthermore, it contributes to economic and social sustainability by creating favourable investment opportunities for producers (Rodić et al., 2011). Nevertheless, modern poultry farming faces challenges, including animal health and welfare, poultry house management, production, and human-induced issues, which are critical for sustainability in poultry farming (Gunnarsson et al., 2020; Hafez and Attia, 2020). Poultry farming management is transitioning from human labour to smart systems facilitated by machines (Ren et al., 2020). The application of smart technologies in poultry farming is expected to enable faster and more effective farm and animal monitoring, leading to better-informed decision-making through the evaluation of extensive data (Sharma and Patil, 2018).

Among the various technological tools, robots are emerging as a prominent solution in poultry farming, serving diverse functions such as phenotyping, monitoring, management, and environmental control (Sahoo et al., 2022). Recently, functional robots have been developed in poultry farming that can perform specific tasks – such as collecting floor eggs and dead birds, thus saving labour and facilitating the production (Astill et al., 2020; Wu et al., 2022; Zhao, 2021). However, research on the impact of robots designed for direct contact with animals on animal health and welfare is limited (Dennis et al., 2020; Parajuli et al., 2020). Additionally, robots have shown potential in studying collective and social behaviour through interaction with animals, with robot-animal interaction presenting a promising research area (Gribovskiy et al., 2018). Such studies are inspired by the rapid social attachment mechanism known as filial imprinting observed in young animals (Bolhuis, 1991; McCabe, 2013;

Vallortigara and Versace, 2022). Robots interacting with animals hold a huge potential in the investigation of social behaviour and ethological research because they enable highly standardized, controlled, replicable and reproducible experimental designs. This innovative approach allows to explore complex social dynamics in various species (Gribovskiy et al., 2018; Krause et al., 2011; Romano et al., 2019).

There is growing interest in robotics for poultry farming. Previous work has explored the potential impact of smart technology in the poultry industry, focusing on robotics, advanced sensors, automation technology, AI (Artificial Intelligence), big data analysis, internet of things, and transportation (Abbas, 2022; Park et al., 2022; Ren et al., 2020; Wu et al., 2022). Robot-animal social interactions and the impact of robots on animal welfare and animal behaviour in poultry had limited coverage. Thus, this review primarily focuses on robots developed for poultry farming, robot-animal interactions, and their applications in social interactions. Moreover, it aims to identify potential research directions to enhance poultry welfare and behaviour. After discussing the current challenges in poultry farming (section 2), we describe robots currently used in different tasks in poultry farming (section 3) and then assess robot-animal interactions (section 4). The discussion examines the general limitations of robots in poultry farming and future directions of research.

## 2. Current Challenges in Poultry Farming

The poultry industry has enhanced production via rapid growth rates in broilers and increased egg productivity in layers (Kleyn and Ciacciariello, 2021). Despite these advancements in poultry production, there are still various challenges for the sector in its current state and future development, including disease outbreaks, welfare regulations, and environmental factors that affect production practices (Fernyhough et al., 2020; Hafez and Attia, 2020; Mottet and Tempio, 2017; Penz and Bruno, 2011). In this section, we outline the current challenges of poultry farming in four key areas: animal health and welfare, poultry house management, production, and human-induced issues.

## 2.1. Animal health and welfare

Ensuring animal health is crucial in poultry farms. Poultry diseases pose a significant threat, as some of them have the potential to escalate into pandemics with far-reaching global consequences (Carenzi and Verga, 2009). To mitigate such risks, continuous monitoring of poultry is essential for disease prevention, biosecurity measures, early diagnosis, and timely treatment (Aggrey, 2009; Pearce et al., 2023). Robotic solutions have been developed to detect diseases and monitor mortality in poultry houses (section 3.2). Furthermore, certain robots, contribute to animal health by encouraging animal activity (section 3.5).

Upholding animal welfare (Webster et al., 2005) and "life worth living" (Mellor et al., 2016) while ensuring sustainable production practices (Yang et al., 2020) is another challenge. To achieve a comprehensive assessment of animal welfare, standardized parameters must be established and accurately monitored (Penz and Bruno, 2011; Wemelsfelder and Mullan, 2014). The evaluation of animal welfare revolves around indicators such as proper nutrition, good health, suitable housing, and appropriate behaviour (Butterworth et al., 2009; Paul et al., 2022). To evaluate animal welfare, it is fundamental to understand the natural behaviour of a poultry species (Abeyesinghe et al., 2021; Ferreira et al., 2020; Nicol, 2020; Putyora et al., 2023), including social behaviour. The robots mentioned in section 4.2 are designed to monitor welfare and support social behaviour by focusing on social interaction among chicks.

## 2.2. Poultry house management

Among livestock systems, poultry systems are considered environmentally friendly, because produce low greenhouse gas emissions (Leinonen and Kyriazakis, 2016; Vries and Boer, 2010) and lower water usage (Gerber et al., 2015; Vaarst et al., 2015). However, they still require special attention to their environmental impact, particularly concerning issues such as ammonia release and nitrate leaching. Environmental impacts on poultry farms arise directly from energy use, housing, and manure management. To enhance environmental sustainability, it is crucial to measure and monitor the level of environmental impacts overall. Improving poultry housing and developing new strategies for manure management have the potential to further improve the environmental sustainability of the poultry industry (Costantini et al., 2021; Leinonen and Kyriazakis, 2016; Vaarst et al., 2015).

Maintaining optimum environmental conditions needs proficient and stable poultry house management at every stage of production (Flora et al., 2022). Environmental factors, including temperature, humidity, ventilation, gas concentration, and lighting, profoundly influence poultry health and performance (ElZanaty, 2014; Sarıca et al., 2018; Zhang et al., 2016). Robots equipped with various sensors can effectively monitor the climatic environmental conditions in the poultry house (see section 3.1).

Another vital aspect is litter management. Contaminants, such as feed residues and faeces, can lead to the proliferation of bacteria in the litter. Accumulation of waste can result in increased ammonia gas levels in the poultry house due to microbial decomposition (Sakamoto et al., 2020). High humidity in the litter also poses a significant problem for flock health and welfare (Sakamoto et al., 2020). Therefore, the litter must be regularly monitored and effectively managed throughout the production cycle (Sakamoto et al., 2020). We discuss robotic solutions for litter management in section 3.4.

*2.3. Production*

Challenges in poultry production encompass ensuring food safety while maintaining low production costs. Expenses related to feed, maintenance, and equipment constitute the fundamental costs, but production losses also significantly impact farming operations (Hafez and Attia, 2020). Identifying low-yielding hens in egg production and closely monitoring their egg-laying behaviour can aid in cost reduction (Aral et al., 2017; Dogan et al., 2018; Wu et al., 2022).

Free range systems in poultry farming are a method of where hens are provided access to outdoor areas for at least part of the day (Miao et al., 2005; Petek and Cavusoglu, 2021). These systems give hens to areas with nests, perches and litter, allowing them greater mobility and opportunities for natural behaviour (Hartcher and Jones, 2017). However, it's important to note that floor egg problems can arise in these systems, leading to reduced production (Oliveira et al., 2019). Collecting eggs from the floor becomes a daily task, which increases labour costs. Additionally, eggs left on the floor can be broken or eaten by birds. Moreover, if the eggs are not collected promptly, they may mix with the litter and manure, elevating the risk of contamination and adversely affecting food quality and safety (Li et al., 2020a; Chai, 2022). Robotics solutions have the potential to streamline egg collection, minimize production losses, and enhance food safety standards in poultry farming, as discussed in Section 3.3.

*2.4. Human-induced issues*

The general duties of breeders in the poultry industry encompass daily care of the animals, health, and welfare control, and monitoring of the poultry house. Additionally, breeders are responsible for the daily egg collection and dispatch in laying hen breeding. However, with the increase in herd size and the adoption of different breeding systems, the observation and management of the herd have become more challenging (Vroegindeweij et al., 2018). Manual observations are labour-intensive, time-consuming, costly, and prone to subjective information (Parajuli et al., 2020). Therefore, the implementation of automatic monitoring equipment and effective use of technology is imperative to achieve efficient monitoring and informed decision-making (Buijs et al., 2018; Buijs et al., 2020; Vroegindeweij et al., 2018).

Furthermore, breeders' increased activities within the poultry house may cause stress in the animals and lead to cross-contamination by carrying disease factors between the birds. Such situations pose risks to occupational health and safety (Ren et al., 2020). The robots discussed in Section 3 offer potential solutions to overcome human-induced problems in poultry houses with their functionalities. Additionally, the comparison between robot-animal interaction and human-animal interaction is presented in section 4.1, shedding light on the potential benefits of using robots to minimize stress and improve animal welfare in poultry farming operations.

## 3. Robots Used in Poultry Farming

The increasing interest in precision and smart agriculture has prompted extensive research into the application of AI and robotics in agricultural production (Usher et al., 2017). Recent advancements in hardware and software, including robots, sensors, 5G networks, and cloud infrastructures, have facilitated the abundant evaluation of data in agriculture. These data are invaluable for assessing and enhancing production during the control and decision-making phases (Park et al., 2022). Robotic systems that operate on farms and assist breeders (Sahoo et al., 2022) are expected to play a more prominent role in the future, equipped with machine capabilities such as perception, reasoning, learning, communication, task planning, execution (Ren et al., 2020).

Robots find application in various agricultural sectors, including planting, livestock, aquaculture, and poultry farming (Sahoo et al., 2022). In the context of poultry farming, both commercial and experimental robots have been developed to perform diverse tasks aimed at enhancing production, reducing the workforce, safeguarding animal health, and improving welfare, making robots increasingly central in this area (Park et al., 2022).

Poultry farming entails several tasks that need constant monitoring, such as identifying sick and deceased animals, monitoring environmental conditions within poultry houses, cleaning, disinfecting litter, and collecting floor eggs. These tasks are laborious and repetitive (Abbas et al., 2022). Robots have proven effective in information detection and production management (Astill et al., 2020). Robots equipped with advanced sensory and decision-making technologies have the potential to efficiently execute designated tasks, enhancing production efficiency (Ren et al., 2020).

Compared to humans, robots offer the promise of superior accuracy, consistency, and efficiency in monitoring birds and their environment (Mamun, 2019; Park et al., 2022). Human observations, in fact, can be subjective, depending on the observer's experience (Ren et al., 2020; Yang et al., 2020), and might be too expensive to be performed constantly. On the contrary, robots equipped with sensors using artificial intelligence and machine learning can continuously gather localized data as they navigate through the poultry house. This sustained real-time data collection can enable the timely detection of diseases, food safety concerns, and indoor environmental conditions through a robust sensor network (Abbas et al., 2022; Park et al., 2022).

Robots can contribute to increased biosecurity and reduced human-animal interactions in poultry houses, as they reduce the need for frequent human intervention (Gittins et al., 2020). Daily inspections are essential to ensure the proper functioning of systems and the well-being of the animals. Breeders must traverse the poultry house multiple times a day to observe the animals and monitor their behaviour and living conditions (Abbas et al., 2022; Parajuli et al., 2020; Park et al., 2022). However, human breeders may inadvertently become disease vectors, transferring pathogens and viruses between houses and cross-infecting flocks, leading to the rapid spread of diseases (Park et al., 2022). By replacing human labour with robots, the potential for human-induced issues is diminished, and biosecurity is improved by reducing human activities in the henhouse. Section 2.4 presents issues related to human-induced problems, while Section 4.1 provides a comparative analysis between human-animal interactions and robot-animal interactions. In contemporary agriculture, robots are increasingly valued for their ability to save labour and enhance management skills (Usher et al., 2017; Zhao, 2021). This section of the paper focuses on the robots used specifically in poultry (Table 1) and discusses them under separate headings according to the tasks they perform, as outlined below: Environmental monitoring (section 3.1); disease control (section 3.2); collecting floor eggs (section 3.3); disinfection and litter management (section 3.4); Encouraging bird activity (section 3.5).

### *3.1. Environmental Monitoring*

To enhance poultry management in both layer and broiler farming, constant monitoring of the poultry house and animals is essential. Environmental monitoring provides valuable data such as farm air quality, temperature, humidity, air velocity, and carbon dioxide levels for poultry management and assessment of animal health and welfare (Park et al., 2022). Real-time data acquisition facilitates informed decision-making, including the maintenance of favourable environmental conditions for optimal production and the early detection of disease outbreaks. Furthermore, these data contribute to improving operational productivity (Astill et al., 2020; Kaur et al., 2021; Olejnik et al., 2022; Wolfert et al., 2017). Therefore, robots equipped with sensors, cameras, and other systems can contribute to the development of the poultry industry (Astill et al., 2020; Zhang et al., 2016). Some commercial robots can now monitor and record the environment, providing valuable data to breeders (Park et al., 2022), to address the challenges outlined in Section 2.

Scout (2023), (formerly known as ChickenBoy, developed by Faromatics, Spain), is a robot that works suspended from the ceiling, about half a meter above the birds. This autonomous robot is equipped with thermal and light cameras, sensors for temperature, humidity, air velocity, CO2, NH3, light, and sound, as well as a laser pointer to stimulate the movement of birds. As indicated in the product specifications, this robot can control the distribution of birds, detect sick and dead birds, and identify wet spots on the litter and drinkers without direct contact. The robot enables early diagnosis of intestinal diseases by monitoring bird faces and provides images for the detection of leg health. The breeders receive updates from the robot via text message or emails.

Poultry Patrol (2019) is produced by a robotics company that designs multi-tasking robots. Per the robot's intended application, the robot, equipped with autonomous and remote-control capabilities, can monitor farms and animals using various types of integrated cameras. It provides early warnings to breeders by identifying sick and deceased birds through remote monitoring and video recording features.

Liu et al. (2016) designed a mobile robot equipped with an intelligent poultry monitoring system. The robot collects environmental parameters and obstacle information related to poultry and transmits this data to the host wirelessly. Subsequently, the host performs data acquisition, processing, display, storage, and remote control. Octopus XO (2021), developed by Octopus Biosafety, is a multi-task robot capable of collecting various environmental data, including temperature, humidity, $CO_2$, ammonia, sound, and light intensity.

Many of the robots designed to provide environmental monitoring and information in poultry have been developed by commercial companies. Often, these companies do not disclose their data, which poses a limitation to the transparency and reproducibility of their findings. Moreover, scientific research and accessible published resources on these poultry robots remain limited. To assess the operational success and effectiveness of these robotic systems, it is crucial to conduct experimental research and share the results. The publication of research findings would facilitate knowledge dissemination and enable fellow researchers and industry professionals to make informed decisions. Robotic systems that have undergone experimental testing and demonstrate the ability to support the traceability of poultry houses are poised to find broader applications in the industry in the future. Therefore, encouraging further research and dissemination of results is essential for the advancement and wider adoption of these technologies in poultry farming.

### *3.2. Disease control*

Pathogenic infections are among the most critical challenges in poultry farming, as they can spread rapidly within the poultry house (section 2.1). Researchers have focused on developing robotic systems to quickly identify sick animals and remove dead birds from the herd (Li, 2016). Equipping robots with sensors for early warning systems allows the monitoring of disease and food safety-related pathogens in birds (Abbas et al., 2022; Park et al., 2022).

Nanny robots (Charoen Pokphand Group) are designed to monitor the body temperature and movements of animals in conventional 3-layer cage systems using thermal cameras. The robot can detect sick and dead chickens by identifying birds with abnormal temperature values and inactivity (Chicken Nannies, 2017). Li (2016) designed a robot to identify sick and dead birds in cages. The robot warns the animals by hitting the cage and detects the movements of the birds using image processing methods. However, the manual operation and hitting action may cause increased stress in birds. Liu et al. (2021) designed a robot with two modes to remove dead chickens from the poultry house. One mode allows for remote control, while the other is autonomous, and the system can work without human intervention. The robotic system includes arms, a conveyor belt, a storage area, and a sweep-in device. Dead chickens are identified using the YOLO v4 algorithm, an object detection network based on deep learning (Bochkovskiy et al., 2020; Redmon et al., 2016). The system exhibits high reliability, with accuracy, precision, and recall rates of 97.5%, 95.24%, and 100%, respectively. However, recognizing dead chickens poses a challenge because the dead birds' shapes are incomplete and look very similar to a healthy chicken in a sitting or lying position. This similarity can impact the accuracy of image classification. To enhance precision and accuracy, the size of the training dataset for the model should be increased and identification errors should be reduced. Li et al. (2022a) developed a robot equipped with a camera and two grippers mounted at the end of a robotic arm, designed to remove dead chickens. The robotic arm (Gen 3, Kinova Inc., Boisbriand, QC, Canada), along with the camera and two grippers (Robotiq 2F-85, Kinova Inc., Boisbriand, QC, Canada) at the end, was securely mounted on a table. The robot underwent testing to assess its ability to grasp and lift dead chickens present on the table under varying light intensities. The robot arm has a maximum payload capacity of 2000 and can move with 7 degrees of freedom, allowing for versatile motion. The success rate of finding and collecting dead chickens was evaluated at different light intensities, resulting in rates of 53.3%, 80%, 86.7%, 90%, and 90% at 10, 30, 60, 70, and 1000 light intensities, respectively. The robotic arm in question has been specifically engineered for the purpose of retrieving deceased chickens from a stationary table. It is important to note that this robotic arm has not yet undergone testing within the dynamic environment of a live poultry house. Furthermore, research findings reveal a noteworthy observation: a decrease in light intensity has been found to significantly impair the performance of the deep learning model. Specifically, this reduction in illumination adversely affects the model's capacity for object detection, image processing, orientation identification, and, ultimately, its ability to execute the final pick-up performance (Li et al., 2022a). As discussed in Section 3.1, Poultry Patrol (2019) utilizes thermal imaging to monitor the body temperatures of chickens as it moves through the poultry house, enabling the identification of sick and dead birds. Similarly, the autonomous robot Scout (2023) employs an infrared and visible light camera to detect deceased chickens and diseases. Both systems monitor temperature and bird movements to identify sick and deceased animals in caged and cage-free systems. To ensure timely intervention for sick birds, these robots should be equipped with early warning systems.

Most existing robots capable of detecting sick birds lack the functionality for capturing and isolating the affected animals, leaving room for improvement. Furthermore, robots designed for the collection of dead birds are currently suitable for free-range systems. Since light intensities in poultry houses can vary due to factors like infrastructure and bird movements, an effective vision system is essential for these robots to accurately detect dead birds. In addition, in-ground rearing systems, especially equipment such as feeders and waterers may pose obstacles to identifying and collecting dead birds, so the movement flexibility of dead bird collection robots should be increased in closed areas around this equipment. One potential solution is the integration of flexible robotic arms and grippers to improve mobility and accessibility (Laschi et al., 2016; Althoefer, 2018; Frazier et al., 2020). Soft grippers (Fras et al., 2016; Hassan et al., 2019) are well-suited for this task due to their intrinsic safety and ability to naturally conform with the shape of the objects, facilitating the interaction with deformable and delicate objects in unstructured and cluttered environments. Soft tactile sensors could be added too (Mandil et al., 2023; Tomo et al., 2017; Ribeiro et al., 2017), further facilitating the safe handling of objects (Zenha et al., 2021) and possibly their recognition (Kirby et al., 2022) and characterization (Ribeiro et al., 2020), to verify that the picked object is the intended one. However, uncertainties remain regarding the performance of robotic arms for collecting dead birds under challenging poultry house conditions.

Consequently, there is a pressing need to develop and refine robots specifically tailored for the efficient collection of deceased birds.

*3.3. Collecting floor eggs*

The transition from cage system to cage-free systems aim to improve the welfare of laying hens (Ochs et al., 2019; Vroegindeweij et al., 2016) by providing them increased space for movement, perching, dustbathing, and nesting. This transition allows hens to spread their wings and express natural behaviours, ultimately leading to a reduction in confinement-related stress (Bhanja and Bhadauria, 2018; Hartcher and Jones, 2017). However, in cage-free systems hens may lay eggs in areas outside the nest, such as corners of the hen house and dim environments (Li et al., 2022b). While cage-free systems provide hens with various areas such as nests, perches, and litter (Hartcher and Jones, 2017), floor eggs are a common occurrence in these systems and reduce production performance (Oliveira et al., 2019). Automatic egg collection robots have been developed to address this issue, as described below.

Unlike commercial robots, scientific research on the use of robots in poultry farming has primarily focused on addressing the difficulty of collecting floor eggs as described in section 2.3. These robots can also reduce human-induced problems mentioned in section 2.4 by reducing the need for human labour in egg collection. Vroegindeweij et al. (2014b) developed an autonomous robot, PoultryBot, for collecting floor eggs in poultry houses. The robot, equipped with a spiral spring on the front for egg collection, successfully collected over 95% of the eggs (Vroegindeweij et al., 2014b). This robot can drive autonomously for more than 3000 m in a commercial poultry house and collect 46% of 300 eggs. A collection failure occurred in approximately 37% of eggs (Vroegindeweij et al., 2018). The researchers suggested that by improving navigation, obstacle handling and control algorithms, the robot could be used in commercial poultry houses and dense animal environments in the future (Vroegindeweij et al., 2018).

Chang et al. (2020) designed a mobile egg collection robot using a computer vision-based platform that can recognize white and brown eggs in free-range farms. The robot moves toward the eggs with visual tracking, collects them, and stores them in its chamber. In experimental tests, the robot collected between 60% and 88% of the eggs on flat and surrounded floors. Additionally, the robot could collect 8 eggs in 10 minutes in a 25 m2 area. For the robot to function efficiently, it relies on a flat surface free of objects such as egg-shaped stones within its operational area (Chang et al., 2020). Therefore, performance enhancements are necessary when deploying it in a free-range system.

Joffe and Usher (2017) developed GohBot, an autonomous egg-collecting robot that uses a mechanical arm with a vacuum mirror to collect eggs. In tests, the success rate of egg collection was 91.6%. Li et al. (2021) developed an egg-collecting robot consisting of a deep learning-based egg detector, arm, gripper, and camera. Eggs detected by image processing algorithms are collected using the robot arm and grippers. The robot collected brown and white eggs with a success rate of 92% to 94%.

Overall, egg collection robots developed for use in cage-free systems face general challenges, such as 1) mobility within the poultry house, including localization, navigation, path planning, and obstacle avoidance, 2) detecting eggs, 3) collecting eggs without breaking them, 4) storing eggs and possibly classifying them according to weight and shape.

*3.4. Disinfection and litter management*

The floor of the coop can become contaminated with bird faeces and food residues, leading to air pollution and the proliferation of pathogens. Regular cleaning and disinfection of the house are necessary to maintain animal health (Wu et al., 2022). Robots can be effectively used for smart production and appropriate disinfection in poultry houses (Feng et al., 2021). Proper litter management is essential for poultry farming, and regular litter scraping can help aerate the litter, preventing fermentation and reducing litter moisture (Tibot, 2021). Robots designed for litter scraping can address litter management challenges (section 2.2) and support animal health (section 2.1).

Feng et al. (2021) designed an anti-epidemic (Feng et al., 2021) and disinfection spray (Feng and Wang, 2020) robot for use in poultry houses and farms. Comprising a robot, transport vehicle, sensors, spraying unit, and controller, it can work autonomously and with remote control. The researchers proposed the "Magnet-RFID" path detection navigation method for autonomous movement, which involves the manual installation of magnets and RFID (Radio Frequency Identification) electronic tags in the work area. The robot successfully ensured sufficient drug concentration in various parts of the cages to kill pathogenic microorganisms (Feng et al., 2021). As indicated in the product specifications, Octopus XO (2021), a multi-tasking robot, can autonomously move within the poultry house to scrape the litter and prevent the formation of scabs. It also reduces ammonia formation by providing better litter drying and performs litter cleaning by spraying a disinfectant solution. Spoutnic-NAV, another robot developed by Tibot, aerates the litter through the forks mounted on the back while moving in the poultry house (Tibot, 2021). Similarly, Poultry Patrol (2019) scrapes the litter while moving along the coop floor, promoting ventilation. To work successfully in the poultry house, these robots must have localization, navigation, and mobility capabilities without causing harm to the animals in cage-free systems. Another potential challenge for these robots is achieving effective litter scraping. Although robots have been developed for both caged and cage-free systems for sanitation and spraying, accurately determining the area and required amount of pesticide for spraying in the poultry house remains a difficulty.

*3.5. Enhancing bird activity*

Birds need physical activity to maintain their health and well-being. Inactive or sedentary behaviour for extended periods can lead to health issues in birds (Abbas et al., 2022). With the development of production systems characterized by rapid growth rates in broilers, fast-growing strains are used in commercial breeding (Zuidhof et al., 2014). It has been reported that faster-growing breeds have higher inactivity, behavioural traits are affected by the growth rate, and fast-growing breeds sit more, feed more, and walk less than slow-growing breeds (Dawson et al., 2021; Hartcher and Lum, 2020). As higher activity can reduce litter contact, more active animals have been evaluated to have better feather cleanliness, lower hock burn levels and better leg health (Casey-Trott et al., 2017; Dixon, 2020; Hartcher and Lum, 2020). Robots have been shown to be effective in encouraging movement in broilers, leading to improved bone quality in animals (Hartcher and Jones, 2017; Janczak and Riber, 2015).

The two main approaches used in free-range poultry farms to encourage animal movement are mobile robots and laser pointers. Mobile ground robots that move within the poultry house trigger the animals around them to move as well (Li et al., 2022b; Tibot, 2021; T-Moov, 2022). Alternatively, robots with laser pointers project laser lights onto the floor, encouraging the birds to move (Scout, 2023). Tibot Technologies claim that their commercial autonomous mobile robots, T-Moov and Spoutnic NAV, increase bird activity in the poultry house and mitigate the issue of floor eggs in cage-free systems. Additionally, increased bird activity resulted in higher feed consumption and a natural weight gain of 300 grams per animal. Moreover, active birds required fewer antibiotics to achieve weight gain naturally (Ren et al., 2020; Tibot, 2021; T-Moov, 2022).

The robot Octopus XO serves the dual purpose of litter cleaning while stimulating bird activity with laser pointers (Octopus XO, 2021). Similarly, the robot Scout (2023) has asserted its capability to stimulate animal activity through the utilization of laser pointers. Li et al. (2022b) reported that a ground robot designed to reduce floor eggs also effectively encouraged bird movement. Another study by Yang et al. (2020) found that robots significantly increased the activity of broilers.

While robots have proven to increase activity in birds, ground robots face challenges such as improving autonomous mobility within the poultry house and obstacle avoidance. Addressing these issues is crucial for the successful implementation of ground robots in poultry farming. Additionally, further experimental data are needed to thoroughly evaluate the operational success of commercial robots. Overall, the use of robots to encourage bird activity in poultry farming holds great promise, and further research and development are essential to overcome existing challenges.

## 4. Robots for social interactions

*4.1. Evaluation of robot-animal interactions*

In this section, we discuss robotic research that examines the effects of robot-animal interactions on animal welfare and health (Table 2). Robotic studies have demonstrated that production, animal health, and welfare can be assessed by monitoring animals and poultry farms using cameras and sensors (section 3). A different topic is how robots designed for tasks that involve direct contact with animals affect their health and welfare.

Robots interacting with birds can have a positive effect by increasing their activity levels (see section 3.5). This increased activity can enhance foot health and maintain feather condition by reducing the time spent on the ground (Yang et al., 2020). Nonetheless, there are concerns about the potential stresses arising from robot-animal interactions (Parajuli et al., 2020). As mobile ground robots move, they may inadvertently hit, physically harm or frighten the birds (Dennis et al., 2020). Fear is a natural response to perceived danger, serving as an adaptive mechanism to protect animals from potential harm (Jones, 1996). However, fear-induced responses can lead to panic reactions, piling, smothering, lameness, and bone damage, such as keel bone damage, which are common welfare problems in cage-free systems (Edgar et al., 2016; Widowski and Rentsch, 2022). Animal behaviour studies on the natural behaviour repertoire of a species can help to identify proxies for fear or adaptation to external stimuli (Fraser, 2009; Mench, 1998; Meuser et al., 2021). Animal behaviour studies can encompass various aspects, including behavioural proxies such as avoidance, exploration, activity levels, and panic reactions (Fraser, 2009; Mench, 1998; Meuser et al., 2021); physiological proxies such as corticosterone levels, feather condition, heart rate, lameness and bone damage (Edgar et al., 2013; Jacobs et al., 2023; Mikoni et al., 2023; Sahan et al., 2006; Wilcox et al., 2023) and other measurement such as vocalizations (Li et al., 2020b; Olczak et al., 2023; Siegford et al., 2023). Avoidance behaviour in animals is considered a response to fear (Brantsæter et al., 2016; Christensen et al., 2021; Olsson et al., 2019; Versace et al., 2021). Conversely, exploration signifies the recognition of the environment, a reduced fear response, and the ability to adapt to new stimuli (Agnvall et al., 2014; Jones, 1996). Monitoring these behaviours can help assess animal welfare (Meuser et al., 2021; Vroegindeweij et al., 2014a).

Avoidance distance tests, which can be employed to detect fear and exploration behaviours during interactions with robots (Parajuli et al., 2020; Vroegindeweij et al., 2014a), are essential for understanding how animals adapt to robots, including studying behaviours such as fear responses and restlessness (Li et al., 2022b). Several studies have evaluated avoidance behaviour within the scope of investigating the effects of robot-animal interactions on birds. Dennis et al. (2020) have compared the response of birds to walking humans and mobile robots simulating commercial rearing conditions for broilers over a 6-week period, concluding that those robots could safely operate among the broiler flock. Animal behaviour was evaluated through camera images as the robot moved along a fixed route inside the coop three times a day. The farmer used the same route for daily work in the poultry house. While the broilers' activity increased during the robot's motion, the increase was lower compared to human observations. Startling behaviour in broilers was also found to be low. The study highlighted that early exposure to robots might lead to habituation and reduced fear in animals. Moreover, startling behaviours like flapping and jumping were not observed before 9 days of age, emphasizing the importance of early exposure to minimize physical harm to birds. The researchers also noted that as chicks grow older and their body sizes increase, there are challenges in the operation of the robot due to the reduced available space for movement.

Parajuli et al. (2018) evaluated the avoidance distances of laying hens and broilers of different ages in response to a robot for bird-robot interactions. Avoidance distances tended to decrease for both laying hens and broilers at older ages. Human-robot interactions were also evaluated by measuring avoidance behaviour (Butterworth et al., 2009). The study also involved a comparison between robot-animal interactions and human-animal interactions (Parajuli et al., 2018). The overall avoidance distances observed in human-animal interactions were shorter than in robot-animal interactions, suggesting better interactions between animals and humans. However, all tested birds had daily exposure to humans before the avoidance distance test, while their exposure to a robot occurred only on the test day. This difference in exposure may have influenced the outcomes. Researchers attributed the avoidance behaviour towards the robot to its sound and speed, suggesting that these aspects should be optimized in future robots (Parajuli et al., 2020, 2018).

Li et al. (2022b) evaluated the impact of a ground robot designed to reduce floor eggs on various production indicators, including egg production, feed consumption, and mortality. They also measured common physiological stress indicators (e.g., serum corticosterone concentration), to assess the potential stress induced by the robot on the animals. Additionally, given that ground robots were shown to promote bird activity (see Section 3.5), the research also examined bone quality and nesting behaviour. The experiment comprised three groups: a group with no robot, a group with 1 week of exposure to the robot, and a group with 2 weeks of exposure to the robot. For each flock, the experiment extended over ten weeks, divided into two phases. Phase 1 spanned from weeks 35 to 39, while Phase 2 covered weeks 40 to 43. The results indicated that the robot treatments did not induce stress in the birds and did not have a statistically significant effect on production performance. Furthermore, the study concluded that although ground robots encouraged bird movement, there were no significant differences in bone quality and nesting behaviour (Li et al., 2022b). It's worth noting that robot manipulation periods of 1 week and 2 weeks may not be sufficient to affect production performance, stress responses, and bone quality. Therefore, it may be recommended to evaluate the longer-term effect of robot manipulation and investigate its effects throughout the entire production period.

Yang et al. (2020) investigated the effects of robots interacting with broilers on broiler production performance, gait score, paw quality, plumage cleanliness, and activities. They used a robot manipulated by a remote controller for the research. The use of robots was found to significantly increase broiler activity, and no negative effects were observed on production performance. These studies evaluated robot-animal interactions for automation and precision management applications in poultry production. Early initiation of robot-animal interactions can lead to reduced fear and faster exploratory behaviour, promoting earlier adaptability. Nevertheless, it remains unclear how robotic applications might elicit responses in birds of different ages. Future research needs to determine the optimal usage times of robots in broiler houses, considering the decreased mobility of robots and restricted activities of animals at advanced ages. Additionally, there is a research gap in understanding how visual, auditory, and mobility features contribute to the exploration and adaptation process for animals becoming accustomed to robots. Examining features such as the size, colour, movement speed, and sound of the robot can shed light on this aspect. Moreover, more information is needed on robot design to prevent injuries in case of physical contact with animals.

*4.2. Social Interactions*

High-capacity brooder machines are used in commercial poultry farming to produce chicks. However, the chicks hatched in incubators are reared separately from the mother hens (Edgar et al., 2016), depriving them of natural stimuli that influence their behaviour and social learning (Roden and Wechsler, 1998). This section focuses on hen-chick interaction to understand the learning process in chicks raised with mother hens and the development of robots to promote social learning in the absence of the mother hen.

Chicks raised by mother hens in natural conditions learn from their mothers about behaviour, pecking habits, resting times, and food preferences (Edgar et al., 2016). Studies evaluating hen-chick interactions have shown that chicks reared with mother hens exhibit synchronization in activity and resting behaviours. They spend less time standing, more time feeding, and less time on perches compared to chicks raised without a mother (Roden and Wechsler, 1998). In the absence of a mother, chicks tend to display more fear behaviour and engage in feather pecking, leading to welfare issues like panic reactions and broken bones in animals (Edgar et al., 2016). Feather pecking is a behavioural issue in which hens engaged in the unwanted behaviour of pecking at, plucking, or pulling out the feathers of other hens within the same flock (Nicol, 2019). This behaviour, if left unmanaged, can result in feather damage, skin injuries, and may even escalate to cannibalism (Cronin and Glatz, 2020; EFSA AHAW Panel (EFSA Panel on Animal Health and Animal Welfare), 2023; Göransson et al., 2023; Kaukonen and Valros, 2019; Ozenturk et al., 2022). It is a significant welfare concern in the poultry farming industry and may be improved by social learning mechanisms (Roden and Wechsler, 1998). Social learning mechanisms, based on filial imprinting, play a crucial role in shaping chick behaviours (Nicol, 2015). Filial imprinting refers to the fast-learning process that enables animals to quickly learn the features of objects by mere exposure (without explicit reward) and become attached to them (Vallortigara and Versace, 2022) at the beginning of life. Social attachment is shown for instance by following responses and distress calls upon separation from the imprinting object. Chicks are excellent subjects for studying spontaneous social interactions and social interactions that follow filial imprinting (Rosa-Salva et al., 2021; Versace et al., 2021; Wang et al., 2023) In natural conditions, imprinting enables chicks to interact with their mother hens, but in the absence of the mother, they may easily imprint on artificial objects that do not have naturalistic appearance (Rosa-Salva et al., 2018; Rosa-Salva et al., 2021; Versace et al., 2018; Wang et al., 2023; Wood and Wood, 2015). Hence, the development of robots for social interaction with poultry chicks is facilitated by the fact that chicks can imprint also on non-naturalistic objects. Research on early predispositions (spontaneous preferences apparent at birth or hatching), is uncovering how simple features, such as upward movement (Bliss et al., 2023), and face-like patterns (Rosa Salva et al., 2010) can trigger social attraction in poultry chicks (reviews in Rosa-Salva et al., 2021; Lemaire et al., 2022; Versace et al., 2018). Research has also noted that chicks prefer patterned objects (Mondada et al., 2013), coloured objects (with red and blue being attractive colours) (Ham and Osorio, 2007) and moving objects (Bateson and Jaeckel, 1974; Versace et al., 2019). Auditory stimuli also play a crucial role in the imprinting process (Bolhuis and Van Kampen, 1991).

Robotic research aims to improve collective and social behaviour by allowing robots to interact with chicks and support their social learning and development (de Margerie et al., 2011). Using robots as social partners is a promising novel approach to studying animal behaviour, with the primary goal of imprinting chicks on robots that can facilitate healthy behaviour.

Robotic research has explored interactions between robots and chicks in laboratory settings based on the imprinting mechanism (de Margerie et al., 2011; Gribovskiy et al., 2018; Jolly et al., 2016; Slonina et al., 2021) (Table 2). For instance, Gribovski et al. (2018) designed an autonomous mobile robot called PoulBot to study collective and social behaviour in chickens. The robot, through its filial imprinting mechanism, was successfully recognized as a congener leader by the chicks, and social integration was demonstrated experimentally. Slonina et al. (2021) developed the RoboChick to identify features that promote the filial imprinting mechanism in chicks. The robot's attractiveness was tested using flashing lights and vocalizations for robot-chick interactions, which did not induce fear in chicks and proved suitable for other experiments. Jolly et al. (2016) examined the social learning mechanism in groups of quail chicks exposed/not exposed to a robot. After spending 3 days together, the chicks exhibited more synchronized activities after early exposure to the robot. Moreover, chicks made more distress calls upon separation from the robot. de Mergerie et al. (2011) used an autonomous robot to evaluate early spatial experiences in quail chicks, finding that animals were motivated to interact spatially with the robot.

Considering the large number of chicks raised separately from mother hens in commercial poultry breeding, these laboratory experiments hold significant potential in the development of robots for poultry farming. However, social interaction research using robots-animal studies is still limited. Nevertheless, with advancements in the perceptual abilities, autonomy, and capabilities of robots, they are expected to be widely used in the future. Furthermore, the development of animal behaviour based on robots' social learning mechanisms can contribute to the improvement of animal welfare.

# 5. Current limitations and future directions of research

In this section, we discuss the current general limitations and future directions of research in robots for poultry, focusing on different functions. Compared to the scale of poultry farming, scientific research, open-source and commercial robots are still limited (Abbas et al., 2022). To assess the operational success of these robots, it is essential to conduct experimental research and share data on the effectiveness and reliability of different robotic systems able support an industry that has a pivotal socioeconomic role worldwide.

Overall, robots for poultry farming are still limited in functionality and adaptability, since most robots are designed to perform a single, specific task (Ren et al., 2020; Wu et al., 2022). Hence, research should expand multi-tasking abilities, via sensor integration and advanced technology, including AI (Alatise and Hancke, 2020). For instance, in the context of collecting floor eggs and managing dead birds, robots face several challenges. These include difficulties in accessing different locations within the poultry house, issues with target identification and capture due to factors such as poultry house infrastructure, bird movements, changing light intensities, and secluded areas. Moreover, robots designed to collect floor eggs encounter challenges in collecting eggs without breakage, as well as sorting and storing them (Chang et al., 2020; Li et al., 2022b; Vroegindeweij et al., 2018). To address these challenges, robots should be designed to operate effectively across various environmental conditions and production systems. To this aim, reliability of visual and tactile perception, combined with flexibility and safety of the movements, are particularly important. The development of algorithms related to object detection, localisation, navigation, path planning, and control is essential. Some robots can be operated manually and by remote control (Li, 2016; Yang et al., 2020), however, there is a need for autonomous work, reducing the need for constant manual or remote control. Visuo-tactile perception is crucial for autonomous robotic systems, especially if they have to grasp and manipulate objects (Navarro-Guerrero et al., 2023).

An important issue is the difficulty in avoiding obstacles while navigating within the poultry house (Dennis et al., 2020; Vroegindeweij et al., 2018). During the movement of robots, the unpredictable actions of the surrounding chickens can impact the detection of static obstacles. Simultaneously, the robots need to make necessary evasive maneuvers with respect to the moving chickens. These requirements impose a high demand on the robot's environmental perception capability and real-time path planning when confronting mobility-related challenges (see Section 2) including those associated with production and human-induced factors. It will be crucial to develop obstacle awareness systems to improve navigation and guarantee animal welfare.

Most studies on robotic applications in poultry farming have primarily focused on free-range systems. However, robots that come into direct contact with animals pose a risk of harming animals and, as a result, may operate at a slower pace (Abbas et al., 2022; Wu et al., 2022). To mitigate these risks and challenges, robots should be designed with the ability to avoid and regulate contact. A current solution is non-contact systems, such as those involving robotic arms mounted on the ceiling of the farm. However, these systems can hardly be implemented in existing farms and would require restructuring or building of dedicated facilities, complicating the logistics of implementation. Hardware solutions to reduce the risks of robot harming animals include protective equipment, such as robots built with soft materials.

In some situations, robot-animal contact is necessary. In cases where robots are deployed for the identification of sick birds, an additional capability for capturing and isolating animals has been shown to be viable when employed within an operational poultry farm in conjunction with the detection system (Liu et al., 2021). For enhanced welfare and biosecurity, robots should also incorporate an early warning system to promptly intervene in cases involving sick birds.

Poultry farms differ in size and organisation (Ren et al., 2020). However, most robotic studies are conducted in controlled experimental environments or within small-scale poultry houses (Vroegindeweij et al., 2018; Wu et al., 2022). Further tests and development are needed for large-capacity poultry houses, including tests on the integration of multiple robots to working together efficiently, in accordance with the size of the poultry house.

Effective and safe robot-animal interactions require knowledge of the species-specific needs in terms of social interactions. Research has just started to address these areas, with few studies that identify social learning mechanisms that can improve welfare and health in interactions with robots (Gribovskiy et al., 2018; Mostafavi et al., 2010). Further research is needed to understand how robots can be best integrated in commercial farms, from the point of view of hardware design and functionality.

Robot-animal interactions present opportunities and challenges. Ground-based robotic systems offer a promising avenue for enhancing animal mobility, with potential benefits for chicken welfare. Such benefits include a reduction in litter contact (Dixon, 2020; Hartcher and Lum, 2020), an improvement in bone quality (Hartcher and Jones, 2017; Janczak and Riber, 2015), as well as enhancement in foot health and feather condition (Yang et al., 2020).

At the same time, robots increase birds' energy consumption, with effects that have just started to be investigated. It has been suggested that this activity might o reduce egg nutrient accumulation in laying hens and decrease egg weight (Li et al., 2022b). Future research should target various parameters such as food consumption, egg weight and food conversion rate to assess the effect of robot use on overall yield in commercial poultry farming.

Potential stress arising from interactions between robots and animals, and whether robots pose lower or higher challenges to animals, are object of research. Ground robots, as they move around poultry houses, exhibit the potential to reduce the incidence of startling behaviour compared to human breeders, while simultaneously mitigating the risk of disease transmission within the poultry house (Park et al., 2022). Differences of responses to robots within the life course have not been investigated enough (Parajuli et al., 2018). Remarkably, research has revealed that chickens exhibit a propensity to form attachments to non-naturalistic agents, such as robots (Gribovski et al., 2018; Slonina et al., 2021). Furthermore, early exposure to robots can effectively mitigate fear reactions towards these artificial agents (Dennis et al., 2020).

The use of robots can be costly, especially for small-scale coops operations (Abbas et al., 2022; Mamun, 2019). It is crucial to conduct economic analyses to assess the viability of using robots in poultry houses. Such analyses should consider their potential effects on human labour, animal health, and production output to make informed decisions about investment.

Creating robots tailored for various functions in poultry farming demands a collaborative approach that delves into multiple domains, such as mechanical engineering, software development, data analytics, genetic animal breeding, animal behaviour, and animal welfare (Zhou et al., 2022). This diverse integration of specialized knowledge is essential in designing robotic solutions that precisely address the intricate demands of poultry farming, ensuring optimal performance, operational efficiency and animal well-being.

## 6. Conclusion

The exploration of robotic technology for poultry farming has enormous promise awaiting realization. The current research landscape, though limited, indicates the potential for robots to innovate poultry farming, reducing labour dependency and significantly enhancing management efficiency by aiding in animal and environmental monitoring. However, this potential has only just begun to be tapped. Further research is needed to fully harness the benefits of robotics in supporting efficient production and promoting animal welfare.

As the poultry industry delves deeper into the integration of robotic technology, the focus must emphasize the critical aspect of robot-animal interactions. Achieving effective solutions calls for the fusion of engineering innovation with a comprehensive understanding of animal needs and behaviour, as underscored by recent research work on imprinting and predispositions in poultry chicks (Rosa-Salva et al., 2021; Versace et al., 2018) and in adult chickens (Dennis et al., 2020; Nicol, 2023). One of the challenges ahead involves the need for increased data sharing and open-source development. Addressing these challenges and fostering collaboration is crucial for a comprehensive understanding of animal welfare.

Overall, the integration of robotic technology and innovation with a deeper understanding of animal needs and societal demands presents a transformative opportunity for enhancing both productivity and welfare in poultry farming. To fully realise this potential, increased research, collaboration, and attention to the animal welfare within robotic applications are essential.

## Acknowledgements

This work was supported by International Postdoctoral Research Scholarship Program (No.2219) of the Scientific and Technological Research Council of Turkey, China Scholarship Council (No. 202208060081), Royal Society Leverhulme Trust fellowship (SRF\R1\21000155: Generalisation using sparse data: from animal to artificial systems) and Leverhulme Trust research grant (Generalisation from limited experience: how to solve the problem of induction).

**Table 1**. Robots used in poultry farming

| Poultry issue | Farming task to solve | Desired general features | Robotic solution already available | Reference | Target farm animal | Robots |
|---|---|---|---|---|---|---|
| Poultry House Management (sec. 2.2) | Environmental monitoring (sec. 3.1) | Mobility<br><br>Communication<br><br>Measurement<br>- Temperature<br>- Humidity<br>- Air speed<br>- $CO_2$<br>- $NH_3$<br>- Noise level<br>- Light<br><br>Obtain information<br>- Collecting data<br>- Message or email<br>- Alarm | Scout | Scout, 2023 | Cage-free farm | 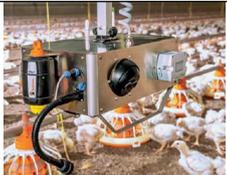 |
| | | | Poultry Patrol | Poultry Patrol, 2019 | Cage-free farm | 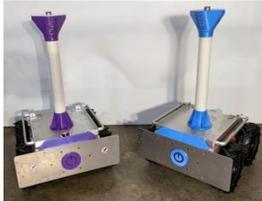 |
| | | | Octopus XO | Octopus XO, 2021 | Cage-free farm | 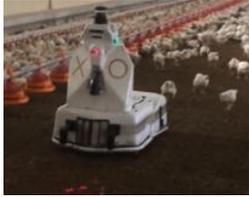 |
| | | | Robot (Liu et al., 2016) | Liu et al., 2016 | Laying hens | |
| Animal Health and Welfare (sec. 2.1) | Disease control (sec. 3.2) | Detecting sick birds<br><br>Detecting dead birds<br>- Temperature<br>- Movement<br><br>Removal dead birds<br>- Vision system<br>- Flexibility<br>- Mobility | Robot (Li, 2016) | Li, 2016 | Laying hens-cage system | 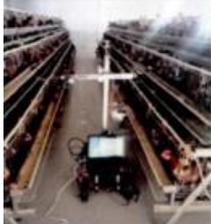 |
| | | | Poultry Patrol | Poultry Patrol, 2019 | Cage-free farm | 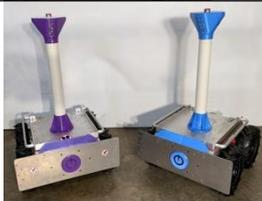 |
| | | | Scout | Scout, 2023 | Cage-free farm | 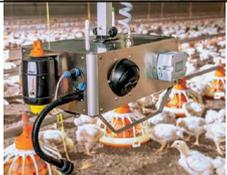 |
| | | | Broiler Removal robot | Liu et al., 2021 | Cage-free farm | 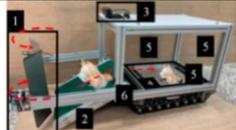 |

| | | | Broiler Removal robot | Li et al., 2022 | Broiler | 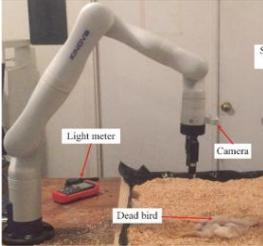 |
|---|---|---|---|---|---|---|
| | | | Chicken Nannies | Chicken Nannies, 2017 | Laying hens-cage system | |
| Production (sec. 2.3) | Collecting floor eggs (sec. 3.3) | Detect and collect floor eggs<br><br>Localisation<br><br>Navigation<br><br>Path planning<br><br>Avoiding obstacles<br><br>Interacting animal<br><br>Sorting<br>- Weight<br>- Size<br>Store | PoultryBot | Vroegindweij et al., 2018 | Cage-free farm, Laying hens | 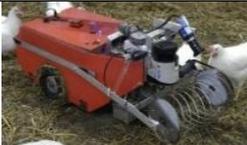 |
| | | | Smart Poultry Robot | Chang et al., 2020 | Cage-free farm, Laying hens | 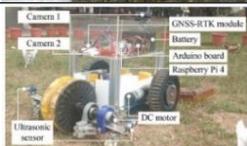 |
| | | | GohBot | Joffee and Usher, 2017 | Cage-free farm, Laying hens | 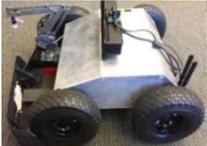 |
| | | | Robotic arm | Li et al., 2021 | Laying hens | 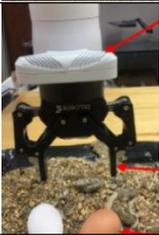 |
| Poultry House Management (sec. 2.2) | Aeration of the litter and disinfection (sec. 3.4) | Localization<br><br>Mobility<br><br>Scraping<br><br>Navigation<br><br>Sanitation-Spraying | Spoutnic Nav | Tibot, 2021 | Cage-free farm | 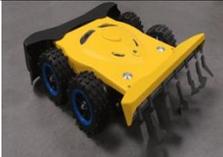 |
| | | | Octopus XO | Octopus XO, 2021 | Cage-free farm | 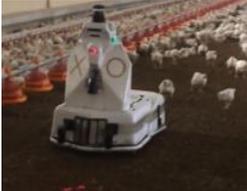 |
| | | | Disinfection Robot | Feng and Wang, 2021 | Cage-free farm | 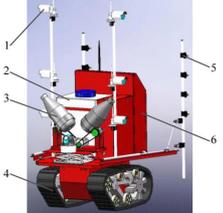 |
| | | | Poultry Patrol | Poultry Patrol, 2019 | Cage-free farm | 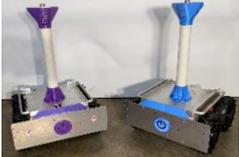 |

| Animal Health and Welfare (sec. 2.1) | Encouraging Bird Activity (sec.3.5) | Trigger the bird's movement<br><br>Autonomous mobility<br><br>Avoiding obstacle | Spoutnic Nav | Tibot, 2021 | Cage-free farm | 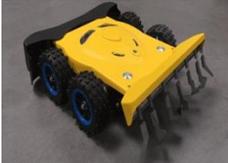 |
|---|---|---|---|---|---|---|
| | | | T-Moov | T-Moov, 2022 | Cage-free farm | 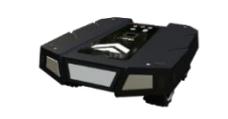 |
| | | | Scout | Scout, 2023 | Cage-free farm | 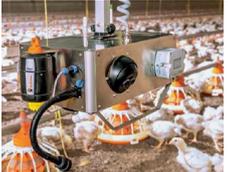 |
| | | | Octopus XO | Octopus XO, 2021 | Cage-free farm | 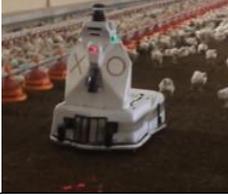 |
| | | | Ground robot | Li et al., 2022b | Cage-free farm | |
| | | | Robotic Vehicle | Yang et al., 2020 | Cage-free farm | |

**Table 2.** Robots for animal interaction and social interaction

| Poultry issue | Farming task to solve | Desired general features | Robotic solution already available | Reference | Target animal | Robots |
|---|---|---|---|---|---|---|
| Animal Health and Welfare (sec. 2.1) | Robot-animal interaction (sec. 4.1) | Reducing animal stress and fear<br>- Observe behaviour<br>- Measure Distance (Avoidance distance tests) | Mobile Robot | Dennis et al.,2020 | Broiler | 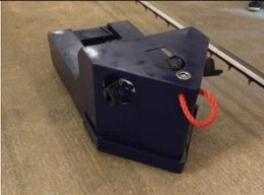 |
| | | | Robotic Vehicle (RV)- Lynx motion | Parajuli et al., 2018 | Hen+broiler | 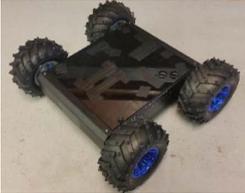 |
| | | | Ground robot | Li et al., 2022 | Broiler | |
| | | | Robotic Vehicle | Yang et al., 2020 | Hen+broiler | |
| | Provide social interaction (sec. 4.2) | Imprinting<br>- Movement<br>- Sound<br>- Colour<br>- Size<br>- Heat source | PoulBot | Gribovskiy et al., 2018 | Chick | 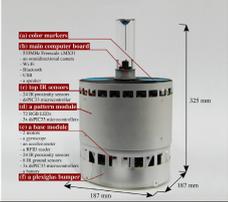 |
| | | | Robo-Chick | Slonina et al.,2021 | Chicks | 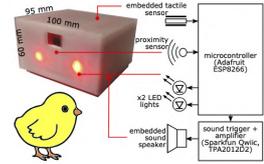 |
| | | | Robot (Jolly et al., 2016) | Jolly et al., 2016 | Chicks | 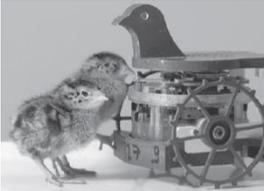 |
| | | | Robot (de Margerie et al., 2011) | de Margerie et al., 2011 | Quail chicks | 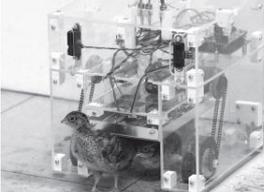 |